\documentclass[10pt,twocolumn,letterpaper]{article}

\usepackage{cvpr}              

\definecolor{cvprblue}{rgb}{0.21,0.49,0.74}
\usepackage[pagebackref,breaklinks,colorlinks,allcolors=cvprblue]{hyperref}
\usepackage{mathrsfs}
\usepackage{multirow}
\def\paperID{1819} 
\def\confName{CVPR}
\def\confYear{2025}

\title{FreePCA: Integrating Consistency Information across Long-short Frames in Training-free Long Video Generation via Principal Component Analysis}

\author{Jiangtong Tan, Hu Yu, Jie Huang, Jie Xiao, Feng Zhao\thanks{Corresponding author.}\\
MoE Key Laboratory of Brain-inspired Intelligent Perception and Cognition,\\
University of Science and Technology of China\\
{\tt\small \{jttan, yuhu520, hj0117, ustchbxj\}@mail.ustc.edu.cn, fzhao956@ustc.edu.cn}
}

\begin{document}
\maketitle

\begin{abstract}
Long video generation involves generating extended videos using models trained on short videos, suffering from distribution shifts due to varying frame counts. It necessitates the use of local information from the original short frames to enhance visual and motion quality, and global information from the entire long frames to ensure appearance consistency. Existing training-free methods struggle to effectively integrate the benefits of both, as appearance and motion in videos are closely coupled, leading to motion inconsistency and visual quality. 
In this paper, we reveal that global and local information can be precisely decoupled into consistent appearance and motion intensity information by applying Principal Component Analysis (PCA), allowing for refined complementary integration of global consistency and local quality. With this insight, we propose \textbf{FreePCA}, a training-free long video generation paradigm based on PCA that simultaneously achieves high consistency and quality. 
Concretely, we decouple consistent appearance and motion intensity features by measuring cosine similarity in the principal component space. Critically, we progressively integrate these features to preserve original quality and ensure smooth transitions, while further enhancing consistency by reusing the mean statistics of the initial noise. Experiments demonstrate that FreePCA can be applied to various video diffusion models without requiring training, leading to substantial improvements. Code is available at \href{https://github.com/JosephTiTan/FreePCA}{https://github.com/JosephTiTan/FreePCA}.
\end{abstract}

\begin{figure*}[t]
  \centering

  \includegraphics[width=0.75\linewidth]{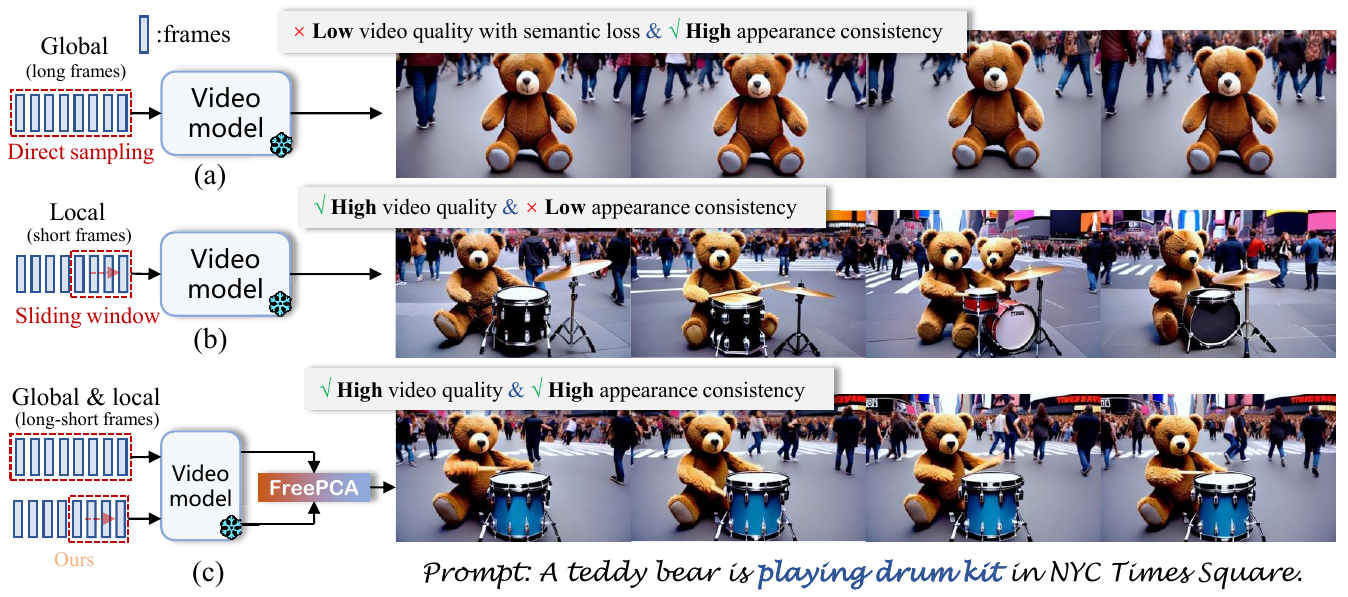}
   \caption{Illustration of different training-free methods for generating long videos. (a) Global aligned method, which inputs the entire video sequence into the model, resulting in lower quality, object loss, and slow motion, but maintains consistency. (b) Local stitched method, which uses a sliding window to extract video segments along the temporal dimension, resulting in poor consistency but retaining the original generation quality. (c) Our FreePCA, which effectively combines the global and local methods via PCA, achieving good consistency while preserving the original generation quality.}
   \label{fig:tes}
\end{figure*}

\section{Introduction}
\label{sec:intro}
As diffusion models have gained prominence in the field of image generation~\cite{ho2020denoising, rombach2022high}, video diffusion generation models~\cite{ho2022video, zhou2022magicvideo, singer2022make} have also begun to develop, leading to various applications such as video editing \cite{wu2023tune, cohen2024slicedit, liu2024video} and motion control \cite{xiao2024video, guo2025sparsectrl, he2024cameractrl, wang2024videocomposer}. 
Previous video diffusion models~\cite{guo2023animatediff, chen2024videocrafter2, wang2023lavie}, trained on large video datasets, have made significant progress in generating high-quality videos. However, training a diffusion model capable of producing long videos remains a resource-intensive task, requiring substantial data collection and annotation~\cite{yin2023nuwa}.

Therefore, researchers have begun to explore methods for generating long videos without training. Directly inputting global noise sequences into models trained on short videos to generate longer videos results in reduced quality, missing objects, and slow motion due to distribution shifts from training data.  As shown in \cref{fig:tes}(a), the concept of ``playing drum kit" is lost, referred to as the global aligned method. To address these issues, some methods~\cite{wang2023gen, qiu2023freenoise}  maintain the original local frame count using sliding windows to align with the training data, then apply fusion techniques to stitch them together into long videos, referred to as the local stitched method. While these methods can maintain video quality, ensuring consistency between windows remains challenging, as seen in \cref{fig:tes}(b).
\cite{lu2024freelong} proposes to generate videos by directly using global noise sequences to enhance video consistency, and fuses it with local attention map in the frequency domain to enhance details. But it relies on features with longer frames to generate the entire video, overlooking the flexibility needed in video generation, which leads to the lack of motion and semantic richness like \cref{fig:tes}(a).
As a summary, how to effectively integrate global information to maintain consistency while ensuring the quality of local information for video generation is significant for this task.

We argue that the local information extracted by sliding windows with original frame count is reasonable for preserving the data distribution of the video generation model and reflects the quality of visual and motion.
While global information reflects overall consistency, as its appearance remains unchanged despite the decline in quality in \cref{fig:tes}(a). Therefore, we aim to find a strong decoupling space to establish an appropriate complementary relationship between global and local information, thereby achieving high consistency and quality in long video generation.
Inspired by methods in the background subtraction task that use Principal Component Analysis (PCA) in the spatial dimension to segment moving foregrounds from  consistent backgrounds \cite{alawode2023learning, goldfarb2014robust, gao2014block}, 
we observe that applying PCA in the temporal dimension allows video sequences to be effectively decoupled into consistent appearance and motion intensity. 
The consistent appearance derived from global information complements local information, enabling effective integration of their advantages (see \cref{{sec:analy}}). This motivates us to integrate this consistency feature generated by the global method into the features of the local method in the principal component space, achieving better consistency and quality, as shown in \cref{fig:tes}(c).

\begin{figure*}[t]
  \centering
  \includegraphics[width=1\linewidth]{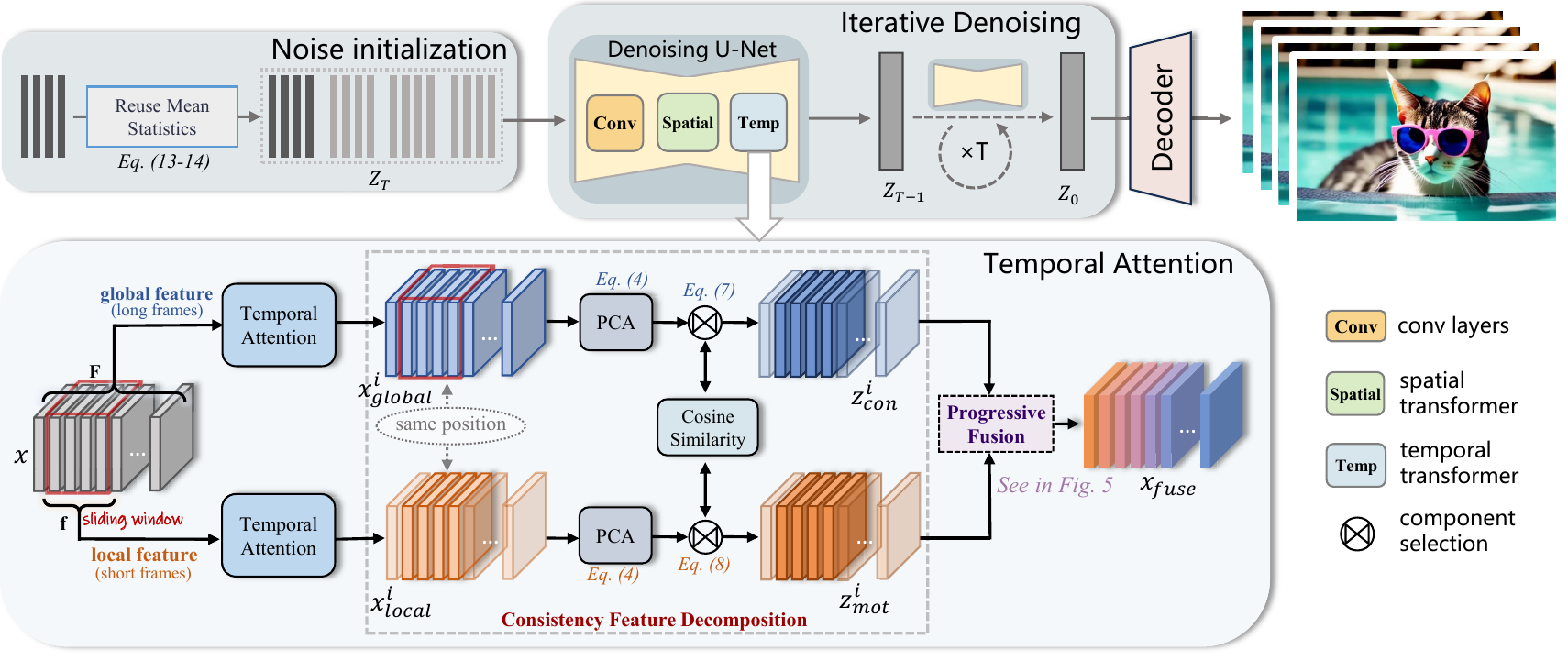}
   \caption{Overview of our method. For noise initialization, we extend short initial noise into long initial noise using a reuse mean statistics approach. In the iterative denoising process, we fuse global and local features in temporal attention through two processes: Consistency Feature Decomposition and Progressive Fusion. For Consistency Feature Decomposition, we crop the global features to the same size as the local features based on their positions. Then, we apply PCA along the temporal dimension to both, using cosine similarity to compare global and local features and decouple them into consistent appearance and motion intensity features. Finally, we perform Progressive Fusion to integrate the global consistent features into the local motion features progressively (see \cref{fig:ppl2} for details).}
   \label{fig:ppl1}
\end{figure*}

In this paper, we propose a training-free method for generating long videos with high consistency and quality via PCA, named FreePCA. Specifically, our approach consists of two steps:
1) \textbf{Consistency Feature Decomposition}. As shown in \cref{fig:ppl1}, we use both long and short frames to generate global and local features, then apply PCA to project them into a shared principal component space, leveraging PCA's strong decoupling capacity to extract consistency features. Given the inherent semantic similarity of global and local features, we employ cosine similarity to compare their components and divide them into consistent appearance and motion intensity features based on their similarity levels, which confirms the fact that appearances generated by both methods are similar but their motion intensities differ. Since the consistent appearance in global features is smoother and more stable than that in local features, we utilize high similarity components of global features as consistency features to complement local features.
2) \textbf{Progressive Fusion}. We gradually increase the proportion of consistency features as the window slides to integrate them into the local features. This maintains the model's flexibility in video generation with the original frame count, preserving consistency and ensuring smooth transitions without compromising video quality.
We also reuse the mean statistics of the initial noise to enhance video consistency further. Experiments show that FreePCA can be applied to various video diffusion models without training, yielding high-quality results while maintaining consistency. Our method supports multi-prompt generation and continuous video generation, demonstrating excellent performance.

In summary, our contributions are as follows:
\begin{itemize}
\item We reveal that PCA can effectively decouple video features into consistent appearance and motion intensity features for the first time, thereby addressing the inconsistency and low quality issues in long video generation.
\item Specifically, we introduce a technique to extract the consistency feature from the entire video sequence's global feature in the principal component space and progressively integrate it into the local feature obtained through the sliding window, which ensures video consistency without compromising video quality.
\item Extensive experiments demonstrate that our method outperforms existing approaches, achieving state-of-the-art results. Moreover, it can be applied to multiple fundamental video diffusion models without extra training.
\end{itemize}


\section{Related work}
\label{sec:related}

\subsection{Text-to-video Diffusion Models}
In recent years, video generation technology has developed rapidly. Initially, this technology relied on networks based on  Variational Autoencoders (VAEs) \cite{mittal2017sync, van2017neural, wu2021godiva} and  Generative Adversarial Networks (GANs) \cite{ deng2019irc, pan2017create}, and it later shifted toward diffusion models \cite{ho2020denoising, ho2022video, zhou2022magicvideo, singer2022make}. Notably, previous work \cite{blattmann2023align, blattmann2023stable, wu2023tune} integrate 3D convolutions and temporal attention to ensure coherent video generation, providing inspiration for subsequent model designs. \cite{guo2023animatediff} fine-tunes diffusion models by adding motion modules, demonstrating strong practical applicability and compatibility with LoRA. \cite{zhang2023show, zeng2024make, he2022latent} employ a latent video diffusion model to compress videos into lower dimensions, enhancing generation efficiency. Building on these, \cite{chen2024videocrafter2} utilizes a video latent diffusion process, leveraging high-quality images to improve visual fidelity. However, due to computational constraints and the scarcity of high-quality video datasets, most existing video diffusion models are trained only on fixed-length short videos, limiting their ability to generate longer content. In this paper, we address this limitation by generating high-quality long videos from short video diffusion models without requiring additional training.

\subsection{Long Video Generation}
Generating long videos has been a challenging task due to the complexities of temporal modeling and resource limitations. Both GAN based \cite{skorokhodov2022stylegan, brooks2022generating, dlpl, urur} and diffusion based \cite{harvey2022flexible, voleti2022mcvd, tseng2023consistent, sstkd, pptformer} methods have proposed solutions for long video generation. Some approaches \cite{yin2023nuwa, zhou2024storydiffusion} require extensive training and entail significant computational costs. Conversely, certain methods \cite{wang2023gen, qiu2023freenoise, lu2024freelong, kim2024fifo} enable long video generation without training, which we categorize into extrapolation and interpolation. Extrapolation methods \cite{wang2023gen, qiu2023freenoise} maintain the original frame generation effect, but often suffer from lower consistency. In contrast, interpolation methods \cite{lu2024freelong} enhance long video sequences with better consistency, though they may compromise visual quality and motion richness compared to the original frames. Our method combines the benefits of both categories, achieving better consistency while preserving both visual quality and motion richness.

\section{Observation and Analysis}
\begin{figure}[t]
  \centering

  \includegraphics[width=1\linewidth]{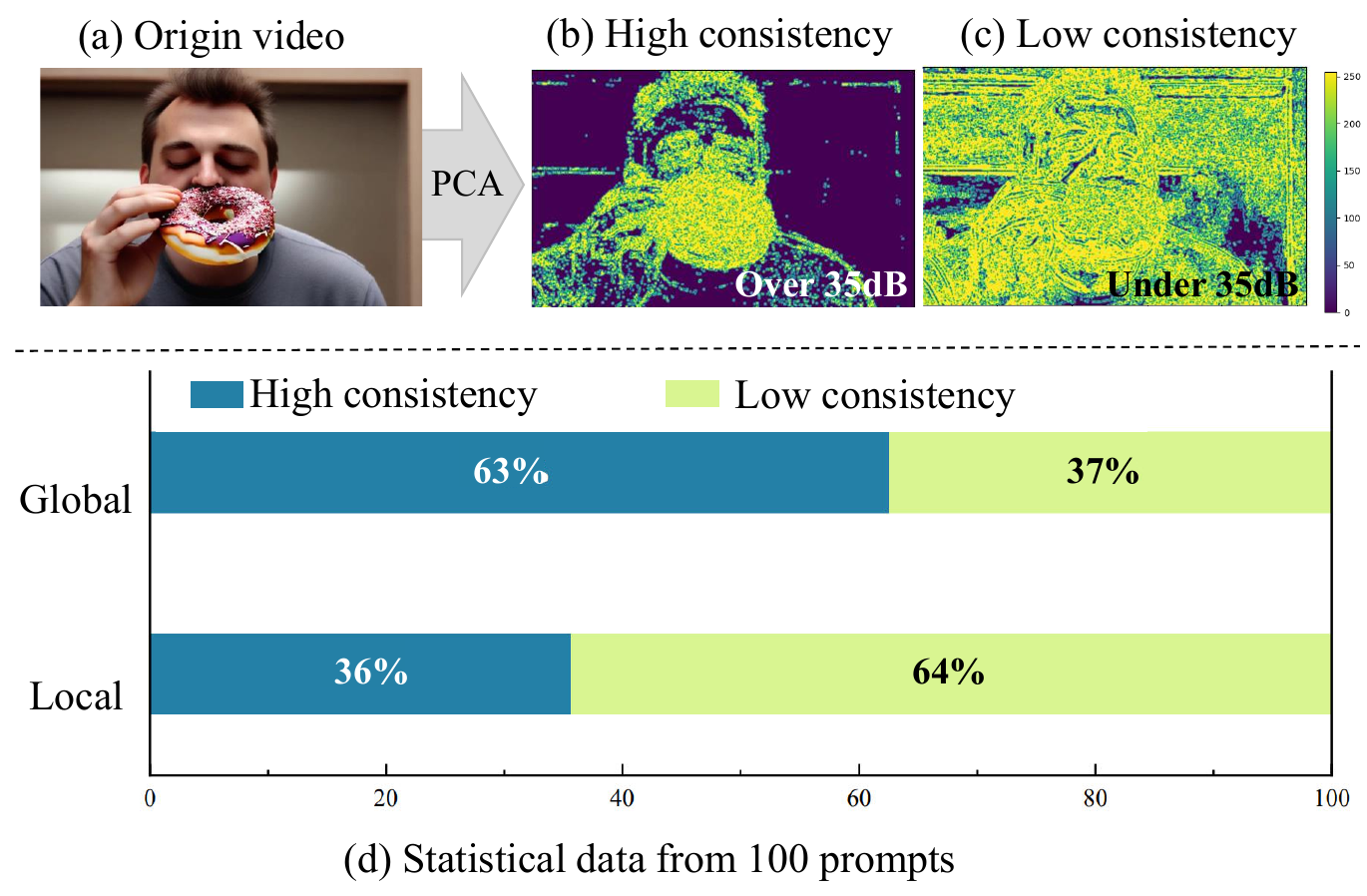}
   \caption{(a-c) Illustration of consistency information extraction after applying PCA to videos and converting each component's information into edge maps. After applying PCA to video (a), some components' information exhibit high consistency like (b), while others show inconsistency like (c). We use PSNR to assess whether each component's information exhibit consistency as the original video, setting a threshold of 35 dB. (d) shows the statistics of 100 videos generated by the global and local methods. We categorize these videos based on whether they have consistency components into two groups: high consistency, and low consistency. It indicates that PCA can separate consistency information and exhibits statistical patterns, reflecting different consistency degree of different methods in the principal component space.}
   \label{fig:o1}
\end{figure}
\label{sec:analy}
In this section, we first introduce the motivation for using PCA and demonstrate that after applying PCA to a video, certain components in the principal component space retain a consistent appearance. We also observe that the proportion of consistency information varies among different long video generation methods statistically. Furthermore, we show how to extract the consistency feature from the video feature in the diffusion model and difference from previous methods.

\textbf{Rationale for using PCA.} Inspired by PCA's information integration capabilities used in video segmentation \cite{alawode2023learning, goldfarb2014robust, gao2014block}, we find that PCA can measure linear correlations between frames in the temporal dimension and enables us to decouple video features into consistent appearance and motion diversity. After applying PCA to the video in the time dimension, we separate the information of each component in the principal component space and then map it back to the original space individually. We find that although there is significant loss of information in each component, some components still retain a consistent appearance attribute. To represent this consistency, we employ Canny edge detection \cite{bao2005canny} on each frame and overlaid all frames. If the edges concentrate in a specific area and display a distinct appearance, it indicates better consistency; conversely, if they are dispersed across a wide area, it suggests poorer consistency. To differentiate this consistency variation, we use PSNR \cite{hore2010image} to measure the distance between the appearance of the video after applying PCA and the original video to determine whether a component exhibits consistency, setting a threshold of 35 dB.
As illustrated in \cref{fig:o1},  after applying PCA, some components reflect a consistent appearance attribute, while others are chaotic and inconsistent.

To analyze the probability of these consistency components from a statistical perspective and their relationship with different generation methods, we applied PCA to videos generated using both long frames (global method) and short frames (local method) with 100 prompts. We categorize the 100 examples into two groups by whether they have consistency components, named high consistency and low consistency. The results indicate that the local method produces a higher number of videos with low consistency compared to the global method statistically. This difference suggests that the degree of consistency of global and local methods can be measured in the principal component space, motivating us to use PCA to address inconsistency issues in local methods and preserve original quality.

\begin{figure}[t]
  \centering
  \includegraphics[width=0.95\linewidth]{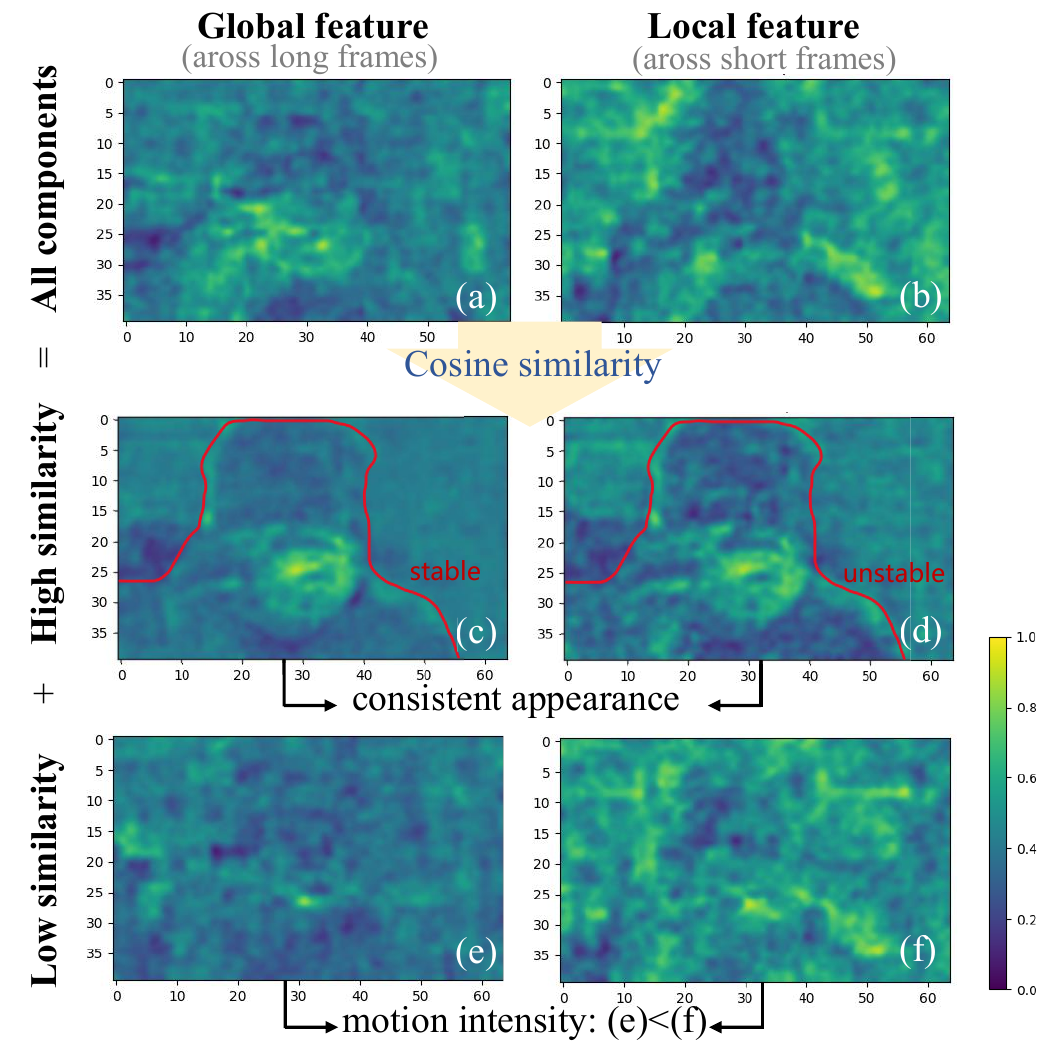}
   \caption{Visualization of consistency features extracted in the principal component space using cosine similarity. After applying both global and local methods in temporal attention and extracting features from the principal component space, we subtract each adjacent frame to exhibit the intensity of changes over time. (a) and (b) show features across all components. (c), (d), (e), and (f) illustrate features selected based on the cosine similarity between global and local features. The distinct character outlines in (c) and (d) indicate that features with high cosine similarity have consistent appearance attributes, while the significant intensity difference between (e) and (f) indicates that features with low cosine similarity exhibit motion intensity attributes.}
   \label{fig:o2}
\end{figure}

Due to the crucial role of temporal attention in long video generation \cite{qiu2023freenoise, lu2024freelong}, we simultaneously employ both global and local methods in temporal attention and extract their features in the principal component space. We find that after comparing the cosine similarity of global and local features across each component, the components with high cosine similarity exhibit consistent appearance attributes, while other components reflect  motion intensity attributes. As visualized in \cref{fig:o2}, we subtract the features of adjacent frames one by one, and the resulting values reflect the intensity of changes with time. After visualizing all components' features, we observe that the local feature (b) exhibits greater change intensity compared to the global feature (a), which is due to the stronger consistency of global features. However, both of them struggle to reveal distinct appearance due to coupled appearance and motion. After applying cosine similarity to select components, features with high cosine similarity display distinct structural appearance, indicating their ability to preserve consistent appearance. Additionally, these features are more stable and smooth in the global feature (c), compensating for the unstable and chaotic shortcomings of the local feature (d). In contrast, features with low cosine similarity cannot display distinct appearances but retain the intensity of changes and have larger values for the local feature (f), preserving more rich motion information than the global feature (e). 
Therefore, in order to achieve precise complementation of their advantages, it is reasonable to integrate features with better consistency (c) into the local features while retaining features (f) with rich motion information to address low quality and inconsistency issues in long video generation. 

\textbf{Difference from previous methods.}
Our method seems to be similar to ~\cite{qiu2023freenoise} and ~\cite{lu2024freelong}, but there are significant differences. We employs PCA that has stronger decoupling capability to decouple video features into consistent appearance and motion diversity and presents clear physical meanings at the feature level. Our method emphasizes how to integrate global consistency with local diversity to achieve superior generation results, an aspect that has not been addressed effectively.

\section{Method}
\label{sec:method}

Based on the analysis, we propose FreePCA, a train-free method for generating long videos with improved consistency and quality using a pre-trained diffusion model grounded in PCA. As shown in \cref{{fig:ppl1}}, the pre-trained model has a U-net structure with convolutional layers, spatial transformers, and temporal transformers, trained on short video data. FreePCA focuses on the temporal transformer and consists of two steps: Consistency Feature Decomposition and Progressive Fusion. We also leverage mean statistics of initial noise to enhance consistency.

\subsection{Consistency Feature Decomposition}
Let the features input to the temporal transformer module be $x\in \mathbb{R}^{(b\times h\times w)\times F\times c}$, where $b, h, w, F, c$ represent batch size, height, width, number of frames, and number of channels, respectively. The pre-trained model is trained on video data with $f$ frames, where $f<F$. After inputting $x$ into the temporal attention using both long and short frames, we represent the global feature obtained by inputting the entire video sequence as $x_{global}\in \mathbb{R}^{(b\times h\times w)\times F\times c}$ and the local feature obtained by using the $i$-th sliding window of size $f$ as $x_{local}^i\in \mathbb{R}^{(b\times h\times w)\times f\times c}$ . Since the number of frames in $x_{global}$ is larger than that in $x_{local}^i$, and to align the features with it, the features of $x_{global}$ are sliced into $x_{global}^i\in \mathbb{R}^{(b\times h\times w)\times f\times c}$according to the positions of the $i$-th sliding window, which can be represented as:
\begin{equation}
  x_{global}^i = Slice^i(Temp(x)), 
  \label{eq:slice1}
\end{equation}
\begin{equation}
  x_{local}^i = Temp(Slice^i(x)),
  \label{eq:slice2}
\end{equation}
where $Temp$ represents temporal attention and $Slice^i$ represents slicing the video sequence with the $i$-th window. Due to the deviation of $x_{global}^i $ from the original distribution, we use the principle of attention entropy \cite{jin2023training} to amplify its query value by a scaling factor $\lambda=\sqrt{\log_fF}$. Details can be seen in \textit{supplementary materials}.

The original PCA process includes data normalization, calculating the covariance matrix, eigenvalue decomposition, and selecting principal components. To simplify the process, we denote the procedure of calculating the transformation matrix $P$ using eigenvalue decomposition before selecting principal components as $\mathbb{T}_{PCA}$. With the frame dimension as the feature dimension for PCA, we project the two features onto a principal component  space using a transformation matrix $P$ from $x_{global}^i$, which can be represented as:
\begin{equation}
  P = \mathbb{T}_{PCA}(x_{global}^i),
  \label{eq:trans1}
\end{equation}
\begin{equation}
  z_{global}^i, z_{local}^i= P\cdot x_{global}^i, P\cdot x_{local}^i,
  \label{eq:trans2}
\end{equation}
where $P\in \mathbb{R}^{f\times f}$. Note that $x_{global}^i$ and $x_{local}^i$ are reshaped from $\mathbb{R}^{(b\times h\times w)\times f\times c}$ to $\mathbb{R}^{f\times (b\times h\times w)\times c}$ before performing the matrix multiplication. And  $z_{global}^i, z_{local}^i\in \mathbb{R}^{f\times (b\times h\times w)\times c}$. 
To extract consistency features, we compare the cosine similarity of each component of global and local features in the principal component space and select the top $k$ most similar components from the $z_{global}^i$, treating their corresponding features as consistency features, and remove original consistency features in $z_{local}^i$, which can be represented as:
\begin{equation}
  {s_{(1)}, s_{(2)},..., s_{(f)}} = CosSim(z_{global}^i, z_{local}^i),
  \label{eq:cos1}
\end{equation}
\begin{equation}
  {n_{(1)}, n_{(2)},..., n_{(f)}} = argsort({s_{(1)}, s_{(2)},..., s_{(f)}}),
  \label{eq:cos2}
\end{equation}
\begin{equation}
  z_{con}^i = z_{global}^i[n_{(1)}, n_{(2)},..., n_{(k)}],
  \label{eq:cos3}
\end{equation}
\begin{equation}
  z_{mot}^i = z_{local}^i[n_{(k+1)}, n_{(k+2)},..., n_{(f)}],
  \label{eq:cos4}
\end{equation}
where $CosSim(\cdot, \cdot)$ represents the computation of cosine similarity $s$ for each of the $f$ components. $argsort(\cdots)$ provides the indices $n$ of the ascending sorted list for cosine similarity. \cref{eq:cos3} and \cref{eq:cos4} indicate ``component selection" to obtain the consistent appearance features $z_{con}^i$ selected from the top $k$ highest cosine similarity components in $z_{global}^i$ and the motion intensity features $z_{mot}^i$ from $z_{local}^i$ after removing the consistency features. 

\begin{figure}[t]
  \centering

  \includegraphics[width=1\linewidth]{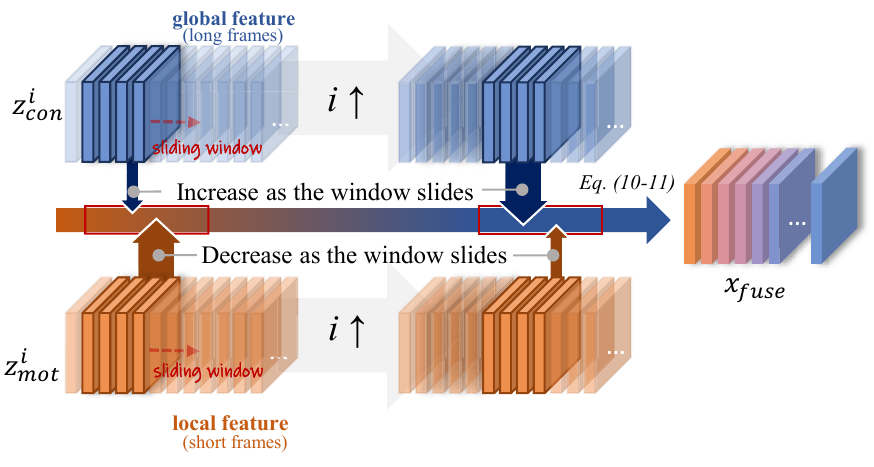}
   \caption{Illustration of Progressive Fusion. As the window slides, the proportion of consistent appearance features from the global features is gradually increased, while the proportion of local motion intensity features is decreased. Finally, the overlapping portions within the window are averaged to obtain the final result.}
   \label{fig:ppl2}
\end{figure}
\subsection{Progressive Fusion}
To avoid affecting the original generation quality of the video model, we gradually add consistency features during the sliding window process. Specifically, we use different values of $k$ for each window to control the proportion of consistency features added. The relationship between $k$ and the $i$-th sliding window can be expressed as:
\begin{equation}
  k = min(i, K_{max}),  i\in \mathbb{N}.
  \label{eq:fu1}
\end{equation}
To preserve the original video generation quality as much as possible, we set a maximum value $K_{max}=3$.
Then, the selected consistency features can be progressively integrated into local features, and finally the transposed matrix $P^{T}$ is used to map them back to the original space:
\begin{equation}
  z_{fuse}^i = Concat(z_{con}^i, z_{mot}^i),
  \label{eq:fu2}
\end{equation}
\begin{equation}
  x_{fuse}^i = P^{T}\cdot z_{fuse}^i,
  \label{eq:fu3}
\end{equation}
where $Concat$ indicates concatenation along the temporal dimension. To get the whole video sequences, we average the values in the overlapping windows and reshape it back to $\mathbb{R}^{(b\times h\times w)\times F\times c}$ to get the final $x_{fuse}.$
Recent research\cite{cao2023masactrl} has suggested that diffusion models initially generate scene layouts and object shapes, followed by fine details in later steps. Therefore, in the 50-step denoising process of DDIM, we use the complete FreePCA for the first 25 steps and employ the local method for the remaining 25 steps to ensure better generation results.

\begin{table*}[t]
    \centering
\caption{Quantitative Comparison. “Direct sampling” indicates directly sampling 64 frames based on short video generation models. The best values are shown in \textbf{bold}.}
    \begin{tabular}{c|c c c|c c c|c}
    \toprule
    \multirow{2}{*}{Methods} & \multicolumn{3}{c|}{Video Consistency} & \multicolumn{3}{c|}{Video Quality} & \multirow{2}{*}{Inference Time (min)}\\
    \cline{2-7}
    & Sub($\uparrow$) & Back($\uparrow$) & Over($\uparrow$)
    & Motion($\uparrow$) & Dynamic($\uparrow$) & Imaging($\uparrow$) \\
    \hline
    Direct sampling & 93.38 & 95.16& 23.52& 92.89& 44.45& 60.02& \textbf{3.6}\\
    FreeNoise \cite{qiu2023freenoise} & 91.98 & 93.86& 25.62& 94.83& 52.77& 63.49& 4.3\\
    FreeLong \cite{lu2024freelong} & 93.77 & 93.79& 24.76& 94.49& 45.83& 63.35& 4.1\\
    \hline
    Ours & \textbf{95.54} & \textbf{95.24}& \textbf{25.69}& \textbf{96.41}& \textbf{59.72}& \textbf{63.70}& 4.7\\
    \bottomrule
  \end{tabular}
  
  \label{tab:base}
\end{table*}
\begin{figure*}[t]
  \centering

  \includegraphics[width=0.9\linewidth]{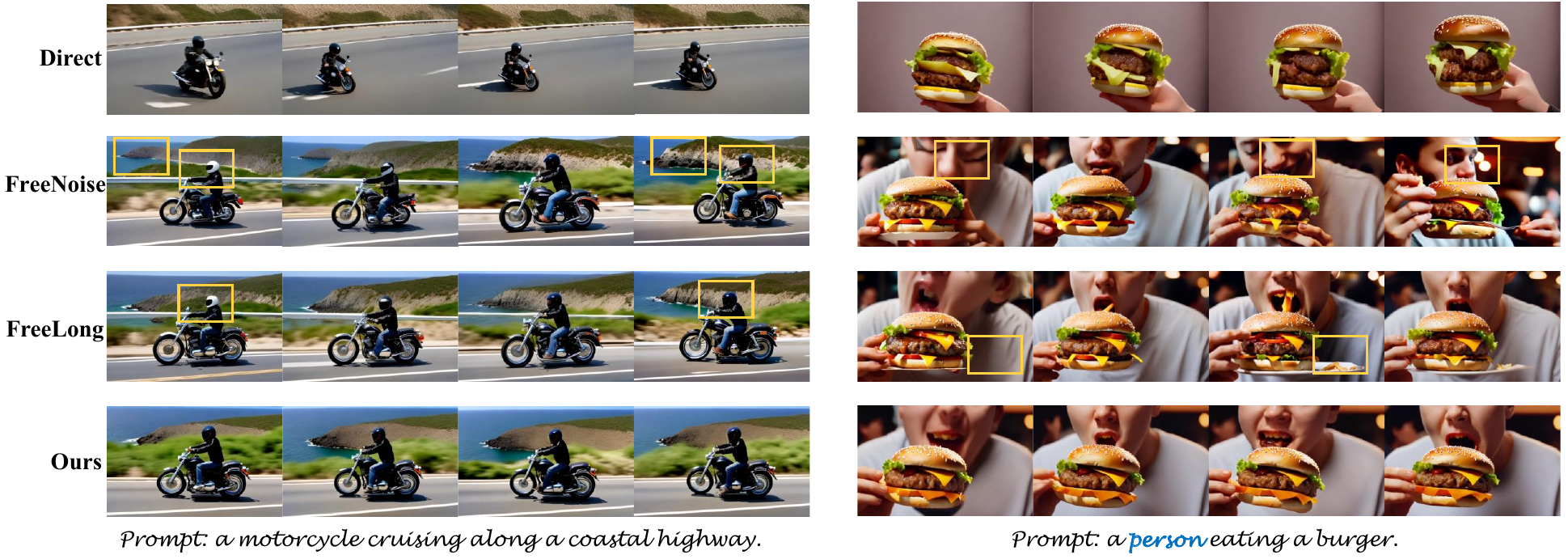}
   \caption{Qualitative comparison using VideoCrafter2 as base model. Direct sampling leads to a loss of detail and semantics, while FreeNoise and FreeLong exhibit inconsistencies. Our method performs best in terms of both quality and consistency.}
   \label{fig:ex1}
\end{figure*}
\begin{figure*}[t]
  \centering

  \includegraphics[width=0.9\linewidth]{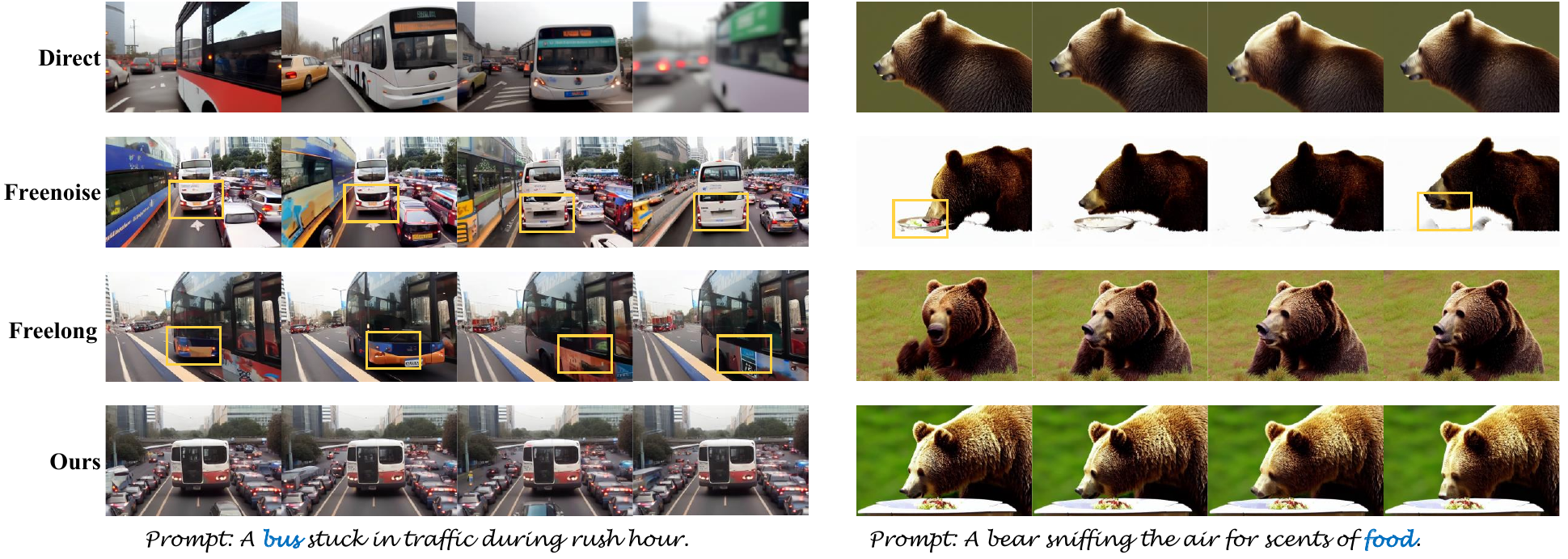}
   \caption{Qualitative comparison using LaVie as base model. Direct sampling results in content blurriness and slow motion, while FreeNoise and FreeLong exhibit inconsistencies and loss of semantics. Our method has the best quality and consistency.}
   \label{fig:ex2}
\end{figure*}
\subsection{Reuse Mean Statistics}
Previous method \cite{qiu2023freenoise} employ noise rescheduling technique to ensure video consistency. However, this approach imposes strict limitations on the input, which hinder the generation of richer scenes. Earlier work \cite{xiao2024video} has suggested that the mean extracted from the video sequence in the temporal dimension can reflect appearance information. Inspired by this, we extract the mean of the first $f$ frames and replace the mean of the subsequent $F-f$ frames of noise. We find that this approach not only maintains the appearance consistency of the video but also enhances the flexibility of video generation, which can be expressed as:
\begin{equation}
  \epsilon_{t} \sim \mathcal{N}(0, 1), t=1,2,..., F,
  \label{eq:mean1}
\end{equation}
\begin{equation}
  \epsilon_{j:j+f}^{\prime} = \epsilon_{j:j+f}-mean(\epsilon_{j:j+f})+mean(\epsilon_{1:f}),
  \label{eq:mean2}
\end{equation}
where $j=nf+1, n\in \mathbb{Z}^{+}$, and $j<F-f.$ Followed \cite{qiu2023freenoise}, we also use the same shuffle way, then all the initial noise can be represented as:
\begin{equation}
  [\epsilon_{1}, \epsilon_{2},..., , \epsilon_{f}, sh(\epsilon_{f+1:2f+1}^{\prime}),..., sh(\epsilon_{j:j+f}^{\prime}),...],
  \label{eq:mean3}
\end{equation}
where $sh(\cdot)$ denotes shuffling the order of frame sequences.

\section{Experiments}
\label{sec:exp}
\subsection{Implementation Details}

\textbf{Setting up}. To validate the effectiveness and generalizability of our method, we apply FreePCA to the publicly available diffusion-based text-to-video models, VideoCrafter2 \cite{chen2024videocrafter2} and LaVie \cite{wang2023lavie}, which are trained on 16-frame video data. Our goal is to enable these models to generate longer videos (i.e., 64 frames) while maintaining the original video generation quality as much as possible. Our method requires no training and can be used directly during the inference phase.

\textbf{Test Prompts}. We use 326 prompts from Vbench \cite{huang2024vbench} to test the effectiveness of our method.

\textbf{Evaluation Metrics}. We employ the metrics available in Vbench\cite{huang2024vbench} to evaluate our approach. We test on two aspects: video consistency and video quality.
For video consistency, we use three metrics:
1) Subject Consistency, which assesses whether objects remain consistent throughout the video by evaluating the similarity of DINO features between frames.
2) Background Consistency, which measures the consistency of the background scene by calculating the similarity of CLIP features between frames.
3) Overall Consistency, which evaluates semantic and stylistic consistency by using ViCLIP features to compute the similarity between frames.
For video quality, we test from two perspectives: motion and appearance, using three metrics:
1) Motion Smoothness, which assesses motion smoothness using the AMT video interpolation model.
2) Dynamic Degree, which estimates the optical flow intensity between consecutive frames using RAFT to determine whether the video is static.
3) Imaging Quality, using the MUSIQ image quality predictor trained on the SPAQ dataset.

\textbf{Baseline}. We compare our method FreePCA with other training-free long video generation methods, including:
1) Direct Sampling, which directly uses a short video model to generate a 64-frame video.
2) FreeNoise \cite{qiu2023freenoise}, which introduce noise rescheduling to maintain consistency between video frames.
3) FreeLong \cite{lu2024freelong}, which blends low-frequency global feature with high-frequency local attention map to improve video quality.
\begin{table}[t]
\small
    \centering
\caption{Result of ablation study.}
    \begin{tabular}{c|c c |c c}
    \toprule
    Config & Sub$\uparrow$  & Over$\uparrow$
    & Motion$\uparrow$ & Imaging$\uparrow$  \\
    \hline
    (1)$K_{max}=1$ & 89.13 & 25.35& 91.46& 60.20\\
    (1)$K_{max}=5$ & 89.40 & 25.14& 91.86& 60.56\\
    \hline
    (2) & 94.19 & 25.40& 95.60& 61.36\\
    (3) & 94.11 & 25.44& 95.40& 59.86\\
    (4) & 88.98& 25.39& 91.69& 59.10\\
    (5) & 94.21 & 25.18& 96.29& 62.49\\
    (6) & 89.54 & 25.16& 92.23& 60.19\\
    \hline
    Ours($K_{max}=3$) & \textbf{95.54} & \textbf{25.69}& \textbf{96.41}& \textbf{63.70}\\
    \bottomrule
  \end{tabular}
  
  \label{tab:abl}
\end{table}
\subsection{Comparison with the Baseline}
Table \ref{tab:base} presents the quantitative results. Directly generating long videos faces issues with domain generalization, leading to a decline in both appearance and motion quality, although its consistency remains relatively acceptable. It also has the worst overall consistency due to its semantic accuracy. FreeNoise does not exhibit significant declines in quality metrics due to sliding windows, but its consistency worsens. FreeLong struggles to further improve quality due to its simplistic frequency fusion way. In contrast, our FreePCA not only achieves superior video quality but also maintains the best consistency due to the use of PCA and the progressive fusion approach. Additionally, we test the inference time of our method on an NVIDIA RTX 4090 and find that our approach achieves better generation results with an acceptable increase in inference time. We also present the results under the DiT framework in the \textit{supplementary materials}.

Qualitative comparisons are presented in \cref{fig:ex1} and \cref{fig:ex2}. It is clear that directly generating long videos results in significant quality degradation, including missing objects, slow motion, and lacking details. FreeNoise exhibits poor appearance consistency. While FreeLong slightly improves consistency, inconsistencies and loss of semantics still exist. In contrast, our FreePCA maintains excellent consistency while ensuring high quality in both appearance and motion.
\begin{figure}[t]
  \centering
  \includegraphics[width=1\linewidth]{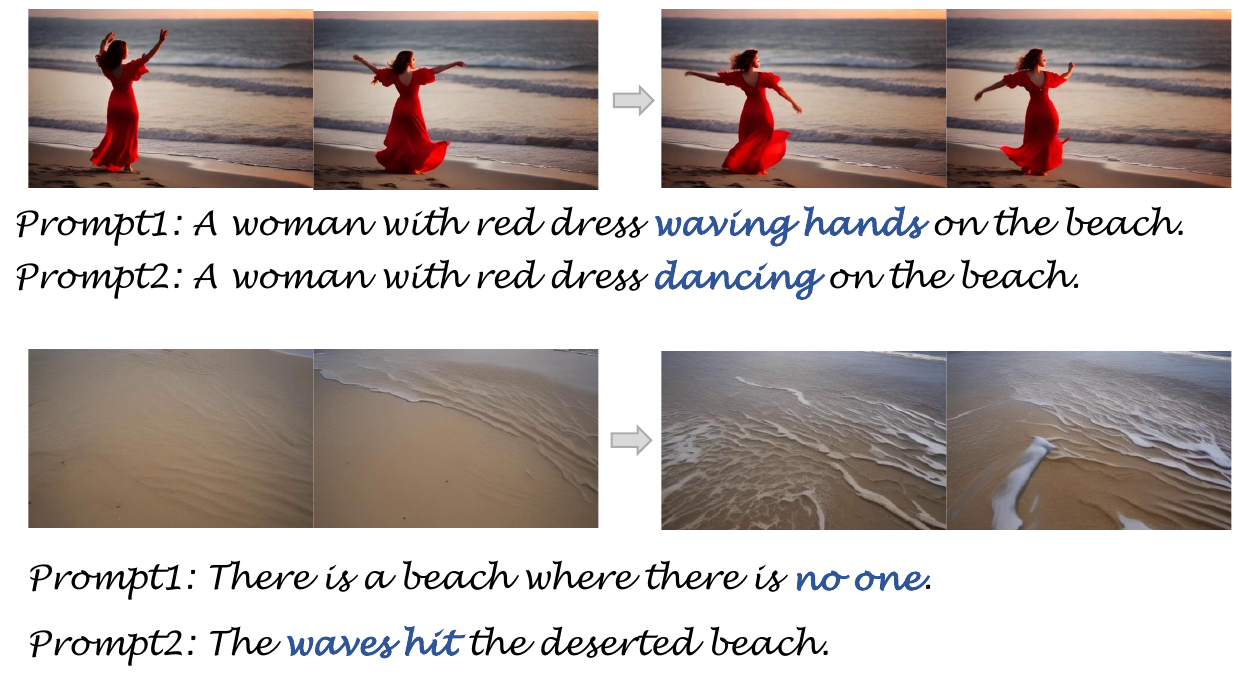}
   \caption{Result of multi-prompt video generation.}
   \label{fig:mu}
\end{figure}
\begin{figure}[t]
  \centering
  \includegraphics[width=1\linewidth]{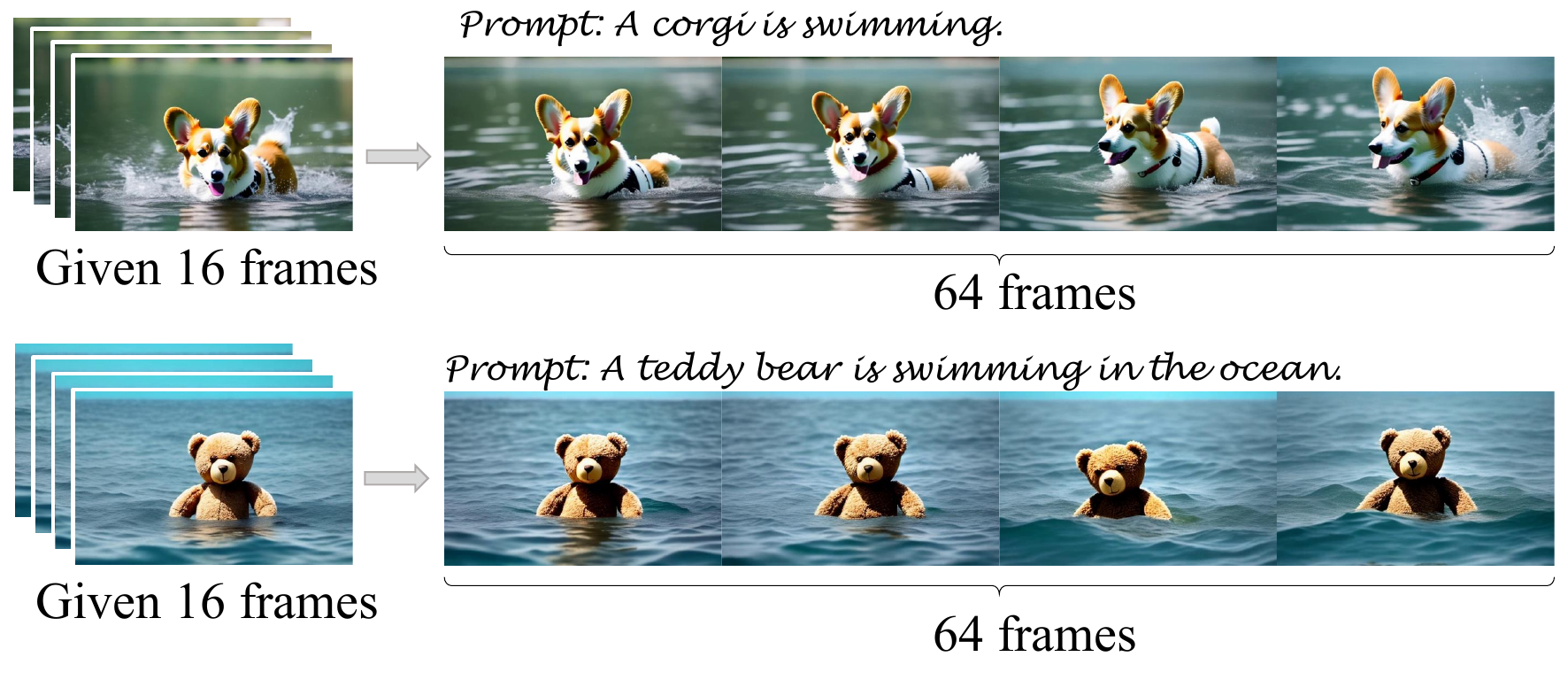}
   \caption{Result of continuing generation based on a given video.}
   \label{fig:long}
\end{figure}
\subsection{Ablation}
 We conduct ablation experiments on:
1) The choice of $K_{max}$.
2) Removing the PCA process.
3) Substituting the cosine similarity selection with random selection.
4) Setting $k=3$ as a fixed value.
5) Replacing the reuse mean statistics with direct reuse.
6) Removing the reuse mean statistics.
The results in Table \ref{tab:abl} show that the overall performance is best when $K_{max}=3$.
It is also evident that optimal results are achieved only when all components of our method are fully utilized. The qualitative comparisons and details are presented in the \textit{supplementary materials}.

\subsection{Enhance Consistency in Other Applications}
Our method not only generates long videos but can also be applied in other scenarios to enhance the consistency of generated videos, such as multi-prompt video generation and continuing video generation based on a given video.

\textbf{Multi-prompt Video Generation}. Our FreePCA can be seamlessly extended to multi-prompt video generation, where different prompts are provided for various video segments. As illustrated in \cref{fig:mu}, our approach enhances the consistency of video generation, maintaining a cohesive appearance even under different prompts.

\textbf{Continuing Video Generation}. This task involves expanding a given short video into a longer video with the same content. By simply using DDIM inversion to the given video before performing FreePCA, we can generate richer, longer video content while keeping the original video unchanged, as illustrated in \cref{fig:long}. The experiments above thoroughly demonstrate the generalizability and practicality of our approach, establishing it as a viable paradigm for maintaining video consistency across various scenarios. Details can be found in the \textit{supplementary materials}.

\section{Conclusion}
\label{sec:con}
In this paper, we introduce FreePCA, a training-free method for generating high quality and consistency long videos from short video diffusion models. Leveraging the strong decoupling capacity of PCA to extract consistency features from video features, we propose Consistency Feature Decomposition, applying cosine similarity after PCA process to identify consistency features. We also design Progressive Fusion, gradually increasing the proportion of consistency features as the window slides to ensure video consistency without sacrificing quality. Additionally, we introduce Reuse Mean Statistics to further enhance consistency. Experiments demonstrate that FreePCA significantly outperforms existing models, achieving high fidelity and consistency, and establishes a training-free paradigm for enhancing consistency in other applications.

\textbf{Acknowledgement.} This work was supported by the Anhui Provincial Natural Science Foundation under Grant 2108085UD12. We acknowledge the support of GPU cluster built by MCC Lab of Information Science and Technology Institution, USTC.

{
    \small
    \bibliographystyle{ieeenat_fullname}
    \bibliography{main}
}


\end{document}